# CoRTEx: Contrastive Learning for Representing Terms via Explanations with Applications on Constructing Biomedical Knowledge Graphs

Huaiyuan Ying[1], Zhengyun Zhao [1], Yang Zhao[2], Sihang Zeng[3], Sheng Yu[1]

[1] Center for Statistical Science, Tsinghua University, Beijing, China

[2] Weiyang College, Tsinghua University, Beijing, China

[3] Department of Biomedical Informatics and Medical Education, University of Washington

Corresponding Author: Sheng Yu, Weiqinglou 209, Center for Statistical Science, Tsinghua University, Beijing, China; Email: syu@tsinghua.edu.cn; Tel: 010-62783842.

# ABSTRACT

**Objective:** Biomedical Knowledge Graphs play a pivotal role in various biomedical research domains. Concurrently, term clustering emerges as a crucial step in constructing these knowledge graphs, aiming to identify synonymous terms. Due to a lack of knowledge, previous contrastive learning models trained with Unified Medical Language System (UMLS) synonyms struggle at clustering difficult terms and do not generalize well beyond UMLS terms. In this work, we leverage the world knowledge from Large Language Models (LLMs) and propose Contrastive Learning for Representing Terms via Explanations (CoRTEx) to enhance term representation and significantly improves term clustering.

**Materials and Methods:** The model training involves generating explanations for a cleaned subset of UMLS terms using ChatGPT. We employ contrastive learning, considering term and explanation embeddings simultaneously, and progressively introduce hard negative samples. Additionally, a ChatGPT-assisted BIRCH algorithm is designed for efficient clustering of a new ontology.

**Results:** We established a clustering test set and a hard negative test set, where our model consistently achieves the highest F1 score. With CoRTEx embeddings and the modified BIRCH algorithm, we grouped 35,580,932 terms from the Biomedical Informatics Ontology System (BIOS) into 22,104,559 clusters with O(N) queries to ChatGPT. Case studies highlight the model's efficacy in handling challenging samples, aided by information from explanations.

**Conclusion**: By aligning terms to their explanations, CoRTEx demonstrates superior accuracy over benchmark models and robustness beyond its training set, and it is suitable for clustering terms for large-scale biomedical ontologies.

# INTRODUCTION

Biomedical Knowledge Graphs (BioMedKGs), exemplified by systems like the Unified Medical Language System (UMLS) [1] and the Biomedical Informatics Ontology System (BIOS) [2], are indispensable in the age of Artificial Intelligence (AI). They play a crucial role in supporting a spectrum of biomedical research areas, including representation learning for Natural Language Processing (NLP) [3-6], drug development [7,8], precision medicine [9,10], and clinical decision support [11-13]. Term clustering, the process of grouping synonymous terms into concepts, emerges as a pivotal step in the construction of modern BioMedKGs [6,14]. Upon mining new terms from extensive corpora, it becomes imperative to discern which terms pertain to the same concept.

Given the vast scale of comprehensive biomedical terminology, neither manual annotations nor pairwise comparison methods prove viable[2,6]. As a result, most existing works turn to encode terms into low-dimensional embeddings and subsequently perform term clustering or entity linking based on embedding similarities [15-17]. The prevailing term embedding models [4-6] predominantly rely on contrastive learning through Transformer-based Pretrained Language Models (PLMs) like Bidirectional Encoder Representations from Transformers (BERT) [20]. This approach generally entails training the model to differentiate between positive samples and negative ones utilizing an encoder architecture [21]. For instance, Yuan et al. [4] introduced a dual contrastive learning approach trained on both synonyms and relations from UMLS. SapBERT [5], on the other hand, concentrated solely on UMLS synonyms but exhibited a more balanced term sampling compared to CODER. An advancement over these models, CODER++ [6], incorporated iterative similarity indexing to sample increasingly challenging terms during the training process, enhancing its ability to distinguish subtle differences compared to its predecessors.

Despite achieving state-of-the-art term clustering performance on UMLS (evaluated on a sampled set), the aforementioned models still grapple with the challenge of limited knowledge injection. Exclusively focusing on aligning synonyms, models like SapBERT and CODER++ tend to overly rely on literal string equivalence, which may result in both false positives and false negatives in the process of clustering. For example, both SapBERT and CODER++ encounter difficulties in distinguishing between terms like "polyaziridine" and "aziridine", which represent distinct concepts despite considerable string overlap. Meanwhile, they struggle to effectively cluster terms that are lexically irrelevant but actually refer to the same concept, such as "gtpase kras g13d" and "kras gly13asp". To address these challenges, models have to transcend simple memorization of synonyms and instead learn a deeper comprehension and knowledge of the associated terms. CODER, though incorporating relations from UMLS into its training, still performs unfavorably as these relations may not convey essential information about the underlying concepts.

Large Language Models (LLMs), exemplified by ChatGPT, have demonstrated remarkable knowledge acquisition across diverse domains including biomedicine [23-25], and there have been numerous works focusing on distilling ChatGPT's extensive knowledge into smaller models tailored for specific tasks [26-28]. In our preliminary investigation, ChatGPT exhibited high accuracy in discerning whether two terms refer to the same concept. However, the direct utilization of this capability is hindered by the quadratic complexity of the term clustering task and the prohibitive costs associated with LLMs. Furthermore, distilling the intricate biomedical knowledge into compact models proves to be a non-trivial challenge [29].

In this work, we introduce Contrastive Learning for Representing Terms via Explanations (CoRTEx). While we employ the same dual encoder contrastive learning framework for its applicability, CoRTEx diverges from prior work by integrating contrastive learning not only for terms but also for their explanations generated by ChatGPT. By prompting ChatGPT to generate concise explanations for each term, we extract the essential knowledge embedded in its vast parameters into natural language. As shown in Figure 1, by aligning the embeddings of the term and its explanation, we inject this knowledge into our term embedding model of a much smaller size. Extensive experiments and case studies demonstrate that our proposed method yields more informative and representative term embeddings, achieving state-of-the-art performance in the term clustering task. Remarkably, the superiority of our term embedding model extends to pure term clustering settings and out-of-distribution data, wherein clustering on unseen terms is based solely on the terms, without their explanations.

As a real-world application, we leverage CoRTEx for term clustering on BIOS, an algorithmically generated BioMedKG where all biomedical terms are automatically mined using machine learning methods. The current version, BIOS V2.2[1], utilizes CODER++ to generate embeddings for each term, and subsequently clusters terms whose embedding similarity exceeds a fixed threshold. The final clustering, representing concepts, is achieved by employing the Ratio Cut algorithm to break down large clusters into smaller ones [30]. However, using a fixed threshold to group tens of

**Figure 1.** Comparison between previous term embedding models and our proposed CoRTEx model.

---

[1] https://bios.idea.edu.cn/

**Figure1**. In conventional encoders (upper), only term synonyms are utilized, which may suffer from insufficient knowledge and lead to poor clustering performances, especially for hard samples. In CoRTEx (bottom), we additionally incorporated term explanations generated by ChatGPT and inject the knowledge into our model via contrastive learning, which contributes to better term embeddings for clustering.

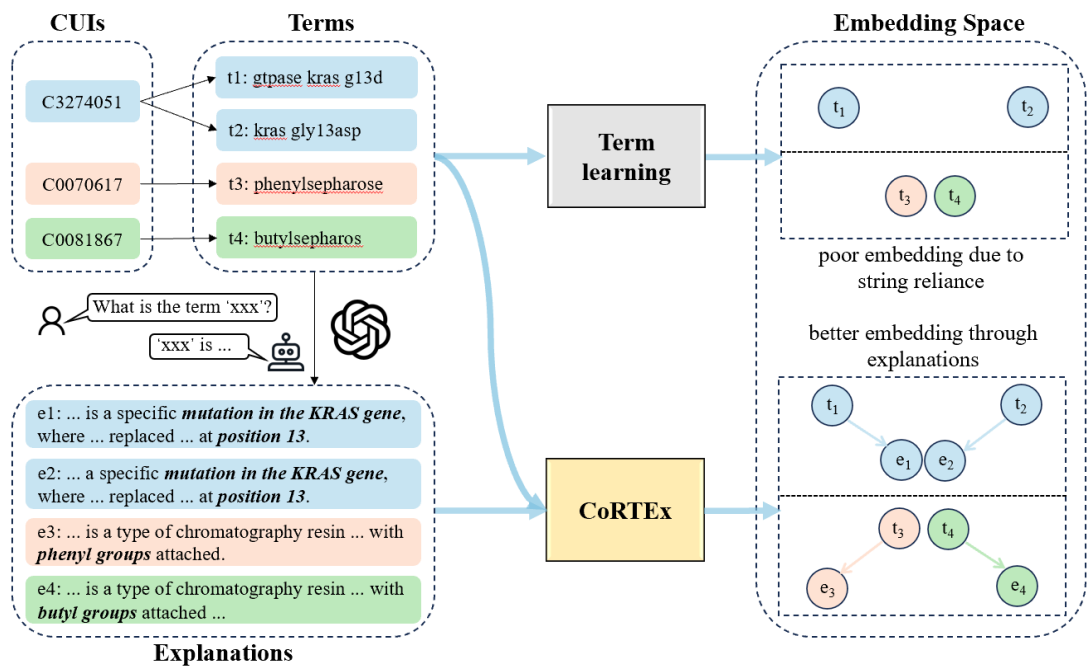

millions of terms by their embeddings trained with contrastive learning leads to suboptimal results. To address this limitation, we turn to ChatGPT and leverage its ability to discern different terms to replace the static threshold. Additionally, we use the BIRCH algorithm [31] instead of Ratio Cut to form the final clusters. Case studies illustrate that the ChatGPT-assisted BIRCH algorithm, on top of CoRTEx-generated term embeddings, significantly outperforms CODER++-based BIOS 2022 in the term clustering task.

In summary, our research makes three key contributions.

1. We introduce the CoRTEx model, which leverages explanations generated by ChatGPT to enhance biomedical term representations, resulting in state-of-the-art performances on term clustering tasks.

2. We demonstrate that the improvement of term embeddings can be extended to previously unseen terms even in the absence of their explanations, showcasing the generalizability of our approach.

3. We propose and employ a ChatGPT-assisted BIRCH algorithm to cluster CoRTEx embeddings of 36 million BIOS terms, successfully organizing them into 22 million meaningful concepts.

## METHODS

In this section, we introduce the CoRTEx model in detail. We utilize a cleaned UMLS subset and augment it with ChatGPT-generated term explanations. Then we apply contrastive learning to train our model, which is initialized from InstructOR [32]. During the training process, we use dynamic negative sampling to enhance the model ability to distinguish hard samples.

### Data construction with ChatGPT-generated explanations

We curated a subset of terms from 11 sources in UMLS 2023AA, including CPT, HPO, MEDLINEPLUS, MSH, MTH, NCI, RXNORM, FMA, MDR, NCBI, and DRUGBANK, These sources are considered reliable for high-quality terms spanning various semantic types. Each term was converted to lowercase, subjected to additional cleaning rules, and terms longer than 5 words were excluded. The refined subset comprises 1,184,475 terms belonging to 823,444 concept unique identifiers (CUIs).

We leverage ChatGPT to generate explanations for these terms. To ensure consistency and prevent context-specific explanations, we employ a stable prompt for querying ChatGPT: "What is the term 'xxx'? Please explain in 50 words as if in a dictionary." The system prompt specifies, "You are a helpful assistant for providing explanations of biomedical terms. You should be regardless of capitalization." Examples will be showcased in the case study section. We assign the same CUIs to the generated explanations as their corresponding terms, and train the model to generate close embeddings for them via contrastive learning. Through the aligning process, we inject explanatory knowledge from ChatGPT into term embeddings. The following sections will not distinguish between terms and explanations, as they are treated in the same way in the training process. The final training set contains 2,444,271 samples.

**Sampling strategies and training procedures**

For model training, we construct positive and negative pairs with our cleaned and augmented data. For each sample $t$ in our dataset, we randomly select $k$ samples with the same CUI as positive samples. Some of the positive samples are easy as they are textually similar. However, there are hard positive samples that the model needs to find their similarity via their explanations. Negative sampling mainly focuses on selecting hard samples, which are crucial to the model performance. A simple yet effective approach is to take the most similar samples with different CUIs from $t$ as hard negative samples. Following CODER++, we periodically update the similarity indices of the terms by their latest embeddings in a vector database to efficiently search for hard negative samples, which makes the training process increasingly challenging to enhance the model.

Similar to previous work, we adopt the Multi-Similarity (MS) loss function [34], which minimizes the similarities of negative samples and maximizes the similarities of positive ones. Specifically, the contrastive learning is conducted on informative negative and positive samples, defined as follows: For an anchor sample $t_i$, let $N_i$ and $P_i$ be the set of all negative and positive samples for this anchor, respectively. Informative negative samples are those having a similarity larger than the minimal positive similarity given a margin:

$$NI_i = \left\{ j \in N_i \mid S_{ij} > \min_{k \in P_i} S_{ik} - \varepsilon \right\},$$

where $S_{ij}$ is the cosine similarity between the embedding of $t_i$ and $t_j$. Similarly, informative positive samples are defined as:

$$PI_i = \left\{ j \in P_i \mid S_{ij} < \max_{k \in N_i} S_{ik} + \varepsilon \right\}.$$

Finally, the loss function is defined as:

$$L = \frac{1}{m} \sum_{i=1}^{m} \left\{ \frac{1}{\alpha} \log \left[ 1 + \sum_{k \in PI_i} e^{-\alpha(S_{ik} - \mu)} \right] + \frac{1}{\beta} \log \left[ 1 + \sum_{k \in NI_i} e^{\beta(S_{ik} - \mu)} \right] \right\},$$

where $m$ is the number of anchor samples in a batch, and $\alpha$, $\beta$, and $\mu$ are hyper parameters.

**Clustering of BIOS terms**

The BIOS ontology comprises 35,880,932 biomedical terms in English, lacking golden labels indicating their cluster membership. Our goal is to encode these terms using CoRTEx and implement an effective clustering method to group them by synonyms. One apparent approach is to set a threshold and group pairs with cosine similarity exceeding the threshold into the same cluster. However, as contrastive learning aims at comparing relativity rather than direct classification, grouping all clusters with a fixed threshold will lead to suboptimal results.

Given the challenge of selecting term-specific thresholds, we turn to ChatGPT for a more intelligent classification. Specifically, for a pair of terms, we query ChatGPT to determine if they refer to the same concept. We use the prompt "Do the terms t₁ and t₂ have roughly the same meaning? Please answer with yes or no only." with the system prompt "You are a helpful assistant for term clustering." This prompt is manually refined to optimize ChatGPT's agreement rate on a randomly selected subset of UMLS, and the employed prompt yields an agreement rate of approximately 0.8.

We revise the BIRCH algorithm [31] to incorporate term embeddings and control querying ChatGPT within $O(N)$ times for clustering $N$ terms. The BIRCH algorithm maintains a tree data structure, where terms contained in each leaf node are deemed as synonyms. We define the center of a node to be the mean of the embeddings under the node. For each term excluding the first one, we traverse layer by layer to find its nearest node by cosine similarity of the embeddings of the term and the node center until reaching the leaf node. At this point, we query ChatGPT to classify whether the term is the same as a randomly selected term in this leaf node. The response from ChatGPT guides whether the term should be added to the existing leaf node or if a new leaf node should be created for it. At any time, if the child number of a non-leaf node surpasses a predefined branching factor, we initiate a split process, ensuring that the two split-out nodes have similar numbers of children. We implement this splitting procedure from the bottom to the root. The algorithm is outlined as follows:

---

***The revised BIRCH algorithm***

**Input:** The terms set $X = \{t_1, t_2, \ldots, t_n\}$, corresponding explanation embeddings $E = \{e_1, \ldots, e_n\}$, branching factor $B$, an initialized rootnode.

**Output:** The constructed BIRCH cluster tree $Tr$

**for** $t_i$ in $X$

---

```
            now_position = rootnode
            While now_position  not_a_leaf_node:
                now_position = find_nearest_child(t_i ,now_position) # by computing cosine similarity
            end While
            # now the t_i would be added to the leaf node whose center is cloest to e_i
            t = random_choose_from(now_position)
            if queryGPT(t,t_i) == True
                now_position.add(t_i)
            else
                now_position.parent.add(set(t_i))
                now_position = now_position.parent
            endif
            update_all_centers_of_nodes()
            While number_of_child (now_position) > B
                if now_position==rootnode
                    initialize new_rootnode
                    newnode1, newnode2 = split_into_two_node(rootnode)
                    newnode1.parent = newnode2.parent = new_rootnode
                    break
                elif now_position  not_a_leaf_node
                    newnode1, newnode2 = split_into_two_node(now_position)
                    newnode1.parent = newnode2.parent = now_position.parent
                endif
                now_position = now_position.parent
            endWhile
    endfor
```

To facilitate parallel execution of ChatGPT queries, we preprocess our entire term set into smaller partitions. Considering the terms as nodes in an empty graph, we initially conducted a fuzzy search to identify the top 100 nearest neighbors for each term using the Faiss vector database [35]. Subsequently, we compute the exact cosine similarity between these neighbors and the term itself. If a pair exhibits a cosine similarity above a threshold of 0.5, an edge is established between the corresponding nodes. Finally, we partition the graph into connected components, forming smaller sets of terms. The revised BIRCH algorithm is then applied to each of these small sets, as terms from different sets are not likely synonyms. This split ensures efficient parallelization.

## RESULTS

**Experimental Settings**

We initialized our model from InstructOR [32], which is essentially a T5-based Transformer encoder [33]. InstructOR is an instruction-finetuned text embedding model designed to handle various tasks by simply adjusting instructions, and we chose it for this versatility to generate embeddings for terms and explanations with a single model. In practice, we use the instruction "represent the biomedical term for identifying synonymous terms" for terms, and "represent the meaning of the biomedical term for retrieval" for explanations.

We set a batch size of 2 on each NVIDIA A800 GPU. One training sample contains one anchor, 15 positive candidates and 15 negative candidates. The model is finetuned on 8 GPUs for 500,000 steps with updating the similarity indices for every 50000 steps. For the hyperparameters in the loss function, we choose the same parameters as in CODER [4].

**Term clustering**

The clustering task seeks to group terms referring to the same concept into the same class according to UMLS. At the sample level, the task is evaluated as binary classification, where a sample consists of a pair of terms and is labeled as positive if they share the same UMLS CUI. All models perform term clustering through embedding similarity: a pair of terms with a cosine similarity exceeding a specified threshold is predicted as positive. For each model, we report performances given by the threshold yielding the highest F1 scores.

*Test Datasets*

We construct two versions of test datasets based on UMLS. The first test dataset (Test 1) evaluates the model's capacity to cluster terms in a general setting. It comprises 10,000 randomly selected UMLS CUIs, each with more than three terms, totaling 57,387 terms. These CUIs are distinct from our training set. To assess the CoRTEx model's understanding of both explanations and terms, we generate explanations and retain the terms that can be recognized by ChatGPT, resulting in a subset comprising only explanations. This subset encompasses 8,249 CUIs and 35,812 terms with their explanations.

Another test set (Test 2) focuses on discerning hard negative samples. In a real-world setting where tens of millions of terms are to be clustered, the proximity between the closest terms referring to different concepts becomes much smaller than when there are only tens of thousands of terms. Therefore, we heighten the difficulty of the test a little adding terms from the training set as potential synonyms for terms in the first test set. For each CUI in Test 1, a random term of the CUI is chosen as the anchor, and 30 nearest neighbors from both the training set and Test 1 are retrieved for comparison. The original CODER [4] is used to identify the 30 nearest neighbors and is excluded from the baselines for a fair comparison. Negative pairs are formed by selecting terms from the 30 neighbors with different CUIs from the anchor. Positive pairs are formed by the anchor with all of its synonyms under the same CUI. This test set comprises 261,740 negative pairs and 47,387 positive pairs. Similarly, we generate an explanation pair set based on this test set, resulting in 251,265 negative pairs and 46,088 positive pairs. In the subsequent sections, we refer to these two sets as the clustering test set and the hard negative test set, each with a subscript of term or explanation, respectively.

*Baseline Models*

**SapBERT [5]:** Self-Alignment Pretraining BERT (SapBERT) is a BERT-based model trained by contrastive learning on UMLS synonyms and uses online hard sample mining for in-batch hard sample selection. The original SapBERT is trained on the entire UMLS and is not suited for evaluation on test sets based on UMLS terms and CUIs. Therefore, we retrained SapBERT on our training set, denoted as SapBERT*, and we denote the original version without the asterisk.

**MedCPT [36]:** Contrastively Pre-trained Transformer for Medicine (MedCPT) is a state-of-the-art biomedical text embedding model initialized from PubMedBERT and utilizes PubMED search logs for training. We include this model due to its superiority in a wide range of biomedical embedding tasks.

**BGE [37]:** BAAI general embedding (BGE) is a state-of-the-art text embedding model in the general domain and has shown remarkable potentials in various knowledge-intensive domains such as biomedicine.

**Table 1.** Main results on the two test sets, with recall (R), precision (P) and F1 scores (F1). The

Term and Explanation columns denote applying corresponding embedding models on terms and explanations, respectively. We highlight the best scores from these models.

| Datasets | Test 1 (Clustering) | | | | | | Test 2 (Hard Negative) | | | | | |
|---|---|---|---|---|---|---|---|---|---|---|---|---|
| | Term | | | Explanation | | | Term | | | Explanation | | |
| Metric | R | P | F1 | R | P | F1 | R | P | F1 | R | P | F1 |
| **CoRTEx** | **0.553** | **0.78** | **0.647** | **0.663** | **0.804** | **0.727** | 0.526 | **0.658** | **0.579** | 0.564 | **0.675** | **0.614** |
| SapBERT* | 0.521 | 0.459 | 0.488 | 0.583 | 0.37 | 0.452 | 0.434 | 0.426 | 0.43 | 0.409 | 0.393 | 0.4 |
| MedCPT | 0.4 | 0.488 | 0.44 | 0.349 | 0.359 | 0.354 | **0.529** | 0.35 | 0.421 | 0.42 | 0.403 | 0.414 |
| BGE | 0.285 | 0.453 | 0.35 | 0.437 | 0.494 | 0.464 | 0.423 | 0.339 | 0.376 | **0.569** | 0.365 | 0.445 |

The results are shown in Table 1. On both datasets, CoRTEx outperforms the other models by a large margin, demonstrating that CoRTEx not only performs well in the general clustering task, but also excels in distinguishing hard negative samples. Moreover, even in absence of explanations (corresponding to the Term columns in Table 1), CoRTEx still performs far better than the other models, which partially verifies that the explanation data helps infuse knowledge into the understanding and representation of unseen terms. More supporting evidence can be found in the ablation study in the Discussion section.

**Application to the BIOS terminology with ChatGPT**

*Clustering results*

Applying the ChatGPT-assisted BIRCH algorithm to the BIOS terminology, we successfully clustered 35,589032 terms into 22,104,559 clusters. Among them, 18,497,595 clusters consist of only one term, while 563,523 clusters comprise more than five terms. Meanwhile, the largest cluster contains 438 terms, which pertains to "Reversed-Phase HPLC technology". The single-term clusters often function as subclasses of larger concepts and lack synonymous terms within the ontology.

## DISCUSSION

*Case Study*

We select examples from BIOS and UMLS, shown in Table 2, to understand the rationale underlying our algorithmic improvement and the efficacy of our embedding. The selected terms all appear in the UMLS, which is used as the gold-standard for the label. The first two pairs of positive examples pose challenges due to their dissimilarity in surface text. However, from the explanations from ChatGPT, one can understand that they involve the same genes and mutations, and therefore, they refer to the same concept. This capability is successfully distilled to CoRTEx, resulting in high embedding similarity both in the terms and their explanations. Note that these four terms are not present in the training set, and the CoRTEx model demonstrates the ability to generalize and learn to express a term through its potential explanations. In contrast, CODER++ and the original SapBERT, though having encountered these terms in their training data, fail to comprehend them, leading to lower similarity scores.

The subsequent three negative samples have subtle differences which are effectively captured by ChatGPT-generated explanations. For instance, the explanation of 'Poly' as 'multi' distinguishes between aziridine and polyaziridine. Similarly, the distinctions in prefixes between phenylsepharose and butylsepharose are elucidated. As ChatGPT detects these nuances in the explanations, the CoRTEx term embeddings also exhibit lower similarities than other models. In these five cases, ChatGPT aligned with the UMLS gold standard. However, instances exist where ChatGPT's judgments diverge from the UMLS. For example, when asked about "Do white whales and beluga whales have roughly the same meaning", ChatGPT answers "Yes, because beluga whale is a type of white whale", while UMLS standards dictate that terms cannot be clustered with their hypernyms. Other discrepancies encompass both reasonable judgments and mistakes. Our algorithms currently lack the capacity to rectify these incorrect clustering results.

**Table 2.** Case study from the BIOS and UMLS ontology with comparison of embedding similarity from different models. Key similarities or differences for understanding the classifications are colored in red.

| Term pairs and Classification | Explanations | Embedding Cosine Similarity | | | |
|---|---|---|---|---|---|
| | | CoRTEx_term | CoRTEx_exp | SapBERT | CODER++ |
| gtpase kras g13d; kras gly13asp *Positive* | GTPase KRAS G13D is a specific mutation in the KRAS gene, where the amino acid glycine (G) is replaced by aspartic acid (D) at position 13. KRAS Gly13Asp refers to a specific mutation in the KRAS gene, where the amino acid glycine (Gly) is replaced by aspartic acid (Asp) at position 13. | 0.869 | 0.868 | 0.683 | 0.039 |
| np_004226.3:p.k409q; klf4 p.lys409gln' *Positive* | The term 'np_004226.3:p.k409q' refers to a specific genetic variant in a protein called NP_004226.3. The variant involves a change in the amino acid at position 409, where a lysine (K) is replaced by a glutamine (Q). 'KLF4 p.Lys409Gln is a specific genetic mutation in the KLF4 gene. This mutation replaces the amino acid lysine (K) with glutamine (Gln) at position 409 in the protein. | 0.808 | 0.831 | 0.145 | 0.665 |
| n(6)-methyladenine; n(6)-cyclopentyl-9-methyladenine *Negative* | N(6)-methyladenine is a chemical modification of the DNA base adenine, where a methyl group is attached to the nitrogen atom at position 6. This modification can affect gene expression and DNA repair processes. N(6)-cyclopentyl-9-methyladenine is a chemical compound. It belongs to the class of adenine derivatives and has a cyclopentyl group attached to the N(6) position and a methyl group attached to the 9th position of the adenine ring. | 0.473 | 0.428 | 0.797 | 0.704 |
| polyaziridine; aziridine *Negative* | Polyaziridine is a chemical compound composed of multiple aziridine units. Aziridine is a three-membered ring containing one nitrogen atom and two carbon atoms. Aziridine is a chemical compound consisting of a three-membered ring containing one nitrogen atom and two carbon atoms. It is highly reactive and can be used in organic synthesis. | 0.454 | 0.569 | 0.81 | 0.719 |
| phenylsepharose; butylsepharose *Negative* | Phenylsepharose is a type of chromatography resin used in biochemistry and biotechnology. It consists of a solid support matrix with phenyl groups attached, allowing for hydrophobic interactions with target molecules. Butylsepharose is a type of chromatography resin used in biochemistry and biotechnology. It consists of spherical beads made of cross-linked agarose, with butyl groups attached to the surface. | 0.416 | 0.517 | 0.762 | 0.763 |

*Ablation Study*

We conducted several ablation studies to understand where the model's improvement comes from. These studies involve: (1) utilizing a BERT-based model instead of the T5-based InstructOR as the base model to investigate the influence of the model size (denoted as CoRTEx$_{BERT}$). (2) not using explanations in the training set to assess the impact of aligning terms to explanations (denoted as CoRT~~Ex~~). (3) combining both (1) and (2) (denoted as CoRT~~Ex~~$_{BERT}$). When it reduces to the third ablation, the model is actually CODER++ but trained on our training set instead of the whole UMLS. The evaluation was conducted using the same test sets as in the Results section, and the corresponding scores are detailed in Table 3. The results presented in the table indicate a significant decrease in model performance for each ablation on both test sets. This underscores the importance of including explanations and leveraging larger base models to enhance overall performance.

Surprisingly, considering the two models with BERT as the base model, there is a notable and unexpected drop in scores. We attribute this phenomenon to two primary factors. Firstly, the training procedure involves iteratively selecting hard negative samples based on the current model. In cases where the training data are not sufficiently large, such as ours with 1.2 million terms, a smaller model may struggle to learn genuinely informative knowledge for term embeddings, making it prone to overfitting. Evidentially, both CoRTEx$_{BERT}$ and CoRT~~Ex~~$_{BERT}$ are initialized from CODER, which initially achieved a 0.45 F1 score in Test 1 (Clustering) and 0.4 F1 score in Test 2 (Hard Negative). However, these scores decline rapidly after the training commences. Secondly, the InstructOR model has the capability to assign different embedding transformations for explanations and terms through the prompt. On the other hand, the BERT model lacks the ability to distinguish between prompts and true input. Without the prompts, training a BERT model with contrastive learning on both explanations and terms becomes challenging, potentially leading to confusion within the model and yielding unsatisfactory results.

Upon transitioning to the 1.5B T5-based model CoRTEx, the enhanced base ability accommodates relatively sparse training data and the amalgamation of originally distinct embedding spaces. Additionally, the improvement from not using explanations (CoRT~~Ex~~) to aligning terms with explanations (CoRTEx) is not only significant in the Explanations sets, but also in the pure Term sets, which again supports the hypothesis that the explanations induced the model's ability to interpret terms from deeper knowledge pretrained into the base model. This knowledge does not seem to exist in smaller models like BERT, as evidenced by the low scores of CoRTEx$_{BERT}$.

**Table 3.** Ablation study results on the two test sets.

| Datasets | Test 1 (Clustering) | | | | | | Test 2 (Hard Negative) | | | | | |
|---|---|---|---|---|---|---|---|---|---|---|---|---|
| | **Term** | | | **Explanation** | | | **Term** | | | **Explanation** | | |
| **Score** | R | P | F1 | R | P | F1 | R | P | F1 | R | P | F1 |
| **CoRTEx** | 0.553 | 0.78 | **0.647** | 0.663 | 0.804 | **0.727** | 0.526 | 0.658 | **0.579** | 0.564 | 0.675 | **0.614** |
| **CoRTEx$_{BERT}$** | 0.170 | 0.132 | 0.149 | 0.268 | 0.114 | 0.16 | 0.397 | 0.25 | 0.307 | 0.451 | 0.307 | 0.365 |
| **CoRT~~Ex~~** | 0.443 | 0.712 | 0.546 | 0.445 | 0.723 | 0.551 | 0.435 | 0.662 | 0.525 | 0.45 | 0.598 | 0.514 |
| **CoRT~~Ex~~$_{BERT}$** | 0.250 | 0.145 | 0.183 | 0.091 | 0.053 | 0.067 | 0.383 | 0.249 | 0.302 | 0.432 | 0.246 | 0.314 |

*Limitations and future work*

Based on the experiments and the ablation study, future work may focus on two key aspects. Firstly, if resources permit, adapting a larger base model alongside utilizing the entire UMLS with LLM-generated explanations as training data is recommended. This approach would introduce more information, allowing the model to learn patterns of explanations for specific types of rare terms. The expectation is that this would significantly enhance the model's generalization ability, rectifying current instances of incorrect predictions where the model failed to focus attention on crucial distinctions. Secondly, the ChatGPT-based algorithm may encounter challenges due to the imperfect alignment between ChatGPT judgments and the UMLS standards and may yield suboptimal clustering results. This can be potentially solved by switching to and fine-tuning an open-source LLM. Finally, the BIRCH algorithm relies on the hit@1 accuracy of embedding similarities, which has not achieved a very high level. This again can be solved by using an open-source LLM, as we can query the equivalence of each term to multiple nearest neighbors instead of 1, with lower costs of budget and time.

# CONCLUSION

In this paper, we present CoRTEx, a novel approach that connects the embeddings of explanations generated by LLMs with terms to enhance biomedical term embeddings. The CoRTEx model establishes itself as the state-of-the-art in biomedical term clustering, demonstrating superior performance compared to baseline models by a significant margin. Additionally, we introduce a modified BIRCH algorithm, incorporating ChatGPT and CoRTEx embeddings, to construct novel term clusters for the BIOS ontology. Ablations and case studies conducted underscore the benefits of employing larger language models and harnessing explanation information. Our future research will delve into the exploration of the potential gains from larger base models and more extensive training data. This ongoing investigation aims to further elevate the capabilities and effectiveness of the large language models in the realm of biomedical term embeddings.

## COMPETING INTERESTS

None declared.

## FUNDING

This work was supported by the Natural Science Foundation of Beijing Municipality (Grant No. Z190024) and the Natural Science Foundation of China (Grant No. 12171270).

## DATA AVAILABILITY

Due to the license of UMLS, we cannot publish our training data. The model could be accessed by the url https://github.com/yinghy18/CoRTEx . The clustered BIOS without term ID could be downloaded via https://cloud.tsinghua.edu.cn/f/aff0fe40b78548159f0c/ .

## CONTRIBUTORSHIP STATEMENT

Huaiyuan Ying and Zhengyun Zhao contributed to the writing of the manuscript. Huaiyuan Ying carried out all the main experiments and part of the ablation studies. Zhengyun Zhao and Sihang Zeng contributed to part of the ablation studies. Yang Zhao is responsible for attempting to obtain a more stable ChatGPT prompt for generating explanations. Sheng Yu is responsible for the guidance of the whole study.